\documentclass[conference]{IEEEtran}
\usepackage{times}

\usepackage[numbers]{natbib}
\usepackage{multicol}
\usepackage[bookmarks=true]{hyperref}
\usepackage{amsmath,amssymb}
\usepackage{microtype}
\usepackage{graphicx}
\usepackage{booktabs}
\begin{document}

\title{HiWET: Hierarchical World-Frame End-Effector Tracking for Long-Horizon Humanoid Loco-Manipulation}

\author{
  \authorblockN{
    Zhanxiang Cao$^{1,2}$,
    Liyun Yan$^{1,2}$,
    Yang Zhang$^{1}$,
    Sirui Chen$^{1}$,
    Jianming Ma$^{1,2}$,\\
    Tianyue Zhan$^{1,2}$,
    Shengcheng Fu$^{3,2}$,
    Yufei Jia$^{4}$,
    Cewu Lu$^{1,2}$,
    Yue Gao$^{1,2,\dagger}$
  }
  \authorblockA{
    $^{1}$Shanghai Jiao Tong University \quad
    $^{2}$Shanghai Innovation Institute \quad
    $^{3}$Tongji University \quad
    $^{4}$Tsinghua University\\
    $^{\dagger}$Corresponding author
  }
}


%

\maketitle

\begin{abstract}
  Humanoid loco-manipulation requires executing precise manipulation tasks while maintaining dynamic stability amid base motion and impacts.
  Existing approaches typically formulate commands in body-centric frames, fail to inherently correct cumulative world-frame drift induced by legged locomotion.
  We reformulate the problem as world-frame end-effector tracking and propose HiWET, a hierarchical reinforcement learning framework that decouples global reasoning from dynamic execution.
  The high-level policy generates subgoals that jointly optimize end-effector accuracy and base positioning in the world frame, while the low-level policy executes these commands under stability constraints.
  We introduce a Kinematic Manifold Prior (KMP) that embeds the manipulation manifold into the action space via residual learning, reducing exploration dimensionality and mitigating kinematically invalid behaviors.
  Extensive simulation and ablation studies demonstrate that HiWET achieves precise and stable end-effector tracking in long-horizon world-frame tasks.
  We validate zero-shot sim-to-real transfer of the low-level policy on a physical humanoid, demonstrating stable locomotion under diverse manipulation commands.
  These results indicate that explicit world-frame reasoning combined with hierarchical control provides an effective and scalable solution for long-horizon humanoid loco-manipulation.
\end{abstract}

\IEEEpeerreviewmaketitle

\section{Introduction}

Recent advances in reinforcement learning (RL) and imitation learning have driven rapid progress in humanoid whole-body motion control~\cite{peng2018deepmimic, gu2024humanoid, zhang2025keep, sferrazza2024humanoidbench}.
Motion retargeting addresses the embodiment gap between humans and robots, enabling expressive motion imitation while maintaining dynamic balance~\cite{he2024learning, he2024omnih2o, he2025hover, cheng2024expressive, ji2024exbody2, cao2025learning, yang2025omniretarget, he2025asap}.
Recent works further leverage diffusion models, large-scale motion datasets, and visual perception for more versatile whole-body control~\cite{yin2025visualmimic, luo2025sonic, liao2025beyondmimic, ze2025twist2, zhang2025track, wang2025omni, shao2025langwbc}.
These results demonstrate that learning-based methods can produce stable, expressive whole-body behaviors for humanoid robots.

\begin{figure}[htbp]
  \centering
  \includegraphics[width=\linewidth]{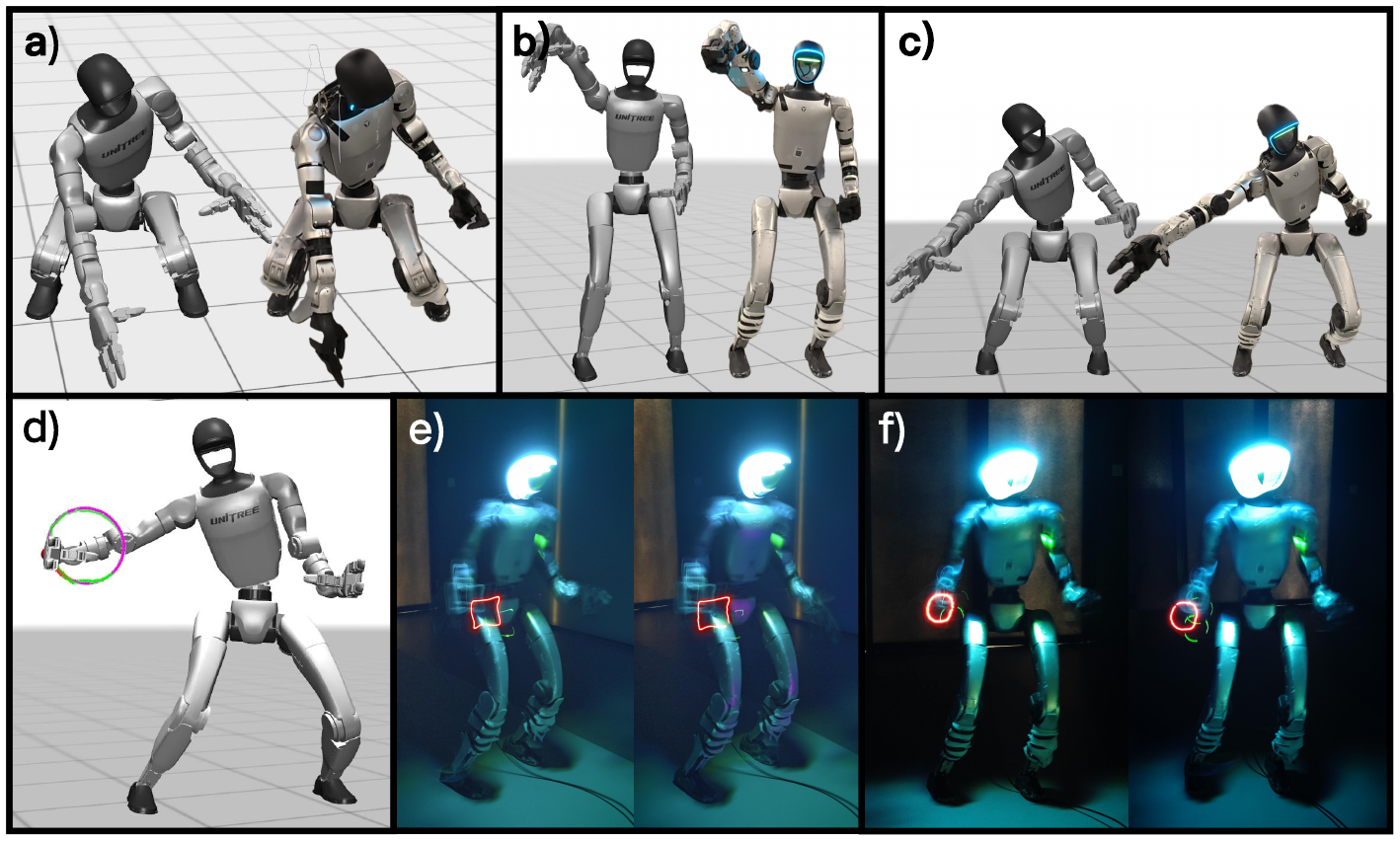}
  \caption{\textbf{HiWET capabilities in simulation and real-world deployment.}
  (a)-(c) Whole-body redundancy exploitation for diverse reaching tasks (left: simulation, right: real robot): (a) lowest and farthest, (b) highest, (c) outermost in semi-squat posture. (d) Sim-to-sim transfer to MuJoCo with world-frame trajectory tracking (red circle: target, green: actual). (e)-(f) Real-world long-exposure experiments: (e) square trajectory, (f) circular trajectory, where red curves are traced by an LED attached to the end-effector.}
  \label{fig:intro}
\end{figure}

Moving beyond motion imitation, command-driven loco-manipulation frameworks enable task-level control through abstract interfaces~\cite{li2025amo, ben2025homie, zhang2025falcon}.
These methods allow operators to specify end-effector poses or joint targets via motion optimization, teleoperation, or structured planners, providing interpretable geometric control with clear task semantics.
Hierarchical architectures further decouple high-level planning from low-level motor control, employing either latent skill interfaces~\cite{zhang2025unleashing, jiang2025wholebodyvla} or explicit trajectory-level commands~\cite{he2025hover, tessler2025maskedmanipulator, portela2025whole, kuang2025skillblender}.
Combined with end-effector stabilization methods~\cite{li2025hold}, these developments have significantly advanced the practical deployment of humanoid loco-manipulation systems.

Despite this progress, fundamental challenges remain.
Whole-body motion imitation optimizes joint-space errors rather than task-space precision, lacking theoretical guarantees for accurate end-effector tracking~\cite{araujo2025retargeting, jiang2025multi} and typically requiring costly dense reference trajectories~\cite{pan2025spider}.
Meanwhile, most command-driven methods operate in body-centric frames, where cumulative base drift and high-frequency oscillations can significantly degrade end-effector precision in the world frame~\cite{li2025clone}.
The tight coupling between upper and lower body dynamics further compounds this challenge: aggressive arm motions can destabilize the gait, while locomotion-induced impacts propagate to the end-effectors~\cite{gu2025humanoid}.
When task trajectories extend beyond the static reachable workspace, the robot must actively transport its base to maintain reachability—a coordination that body-centric formulations do not explicitly address~\cite{sandakalum2022motion}.
We argue that formulating the task as \textit{world-frame end-effector tracking} exposes this geometric coupling, enabling the controller to actively compensate for base disturbances while reshaping the operational workspace through locomotion.

To address these challenges, we propose HiWET, a hierarchical reinforcement learning framework that enables precise world-frame end-effector tracking through explicit upper-lower body coordination.
The high-level \textit{world-frame command policy} reasons in global coordinates, generating structured subgoals—base velocity, body height, and local end-effector targets—to guide the robot toward task objectives while maintaining reachability.
The low-level \textit{whole-body tracking policy} operates at high frequency, translating these subgoals into joint commands while ensuring dynamic stability.
This decomposition separates \textit{where to go} (global planning) from \textit{how to move} (dynamic execution), allowing each policy to specialize in its respective objective.
To accelerate learning and improve tracking precision, we introduce a Kinematic Manifold Prior (KMP) that provides kinematically consistent references for upper-body control, enabling the policy to learn residual corrections rather than absolute joint targets.

Our main contributions are as follows:
\begin{itemize}
  \item We propose a \textbf{world-to-body hierarchical control scheme} that coordinates upper-body manipulation with lower-body locomotion through an explicit spatial interface, enabling world-frame consistency via active base transport and height adjustment.
  \item We introduce an \textbf{efficient Kinematic Manifold Prior (KMP)} integrated within a residual action space, providing high-speed kinematic references that ground the policy in valid manipulation manifolds while preserving dynamic adaptability.
  \item We perform \textbf{extensive validation on a physical humanoid platform}, demonstrating 12.4~mm world-frame tracking error in simulation and robust zero-shot sim-to-real transfer across diverse limb configurations.
\end{itemize}

\section{Related Works}

\subsection{Humanoid Whole-Body Motion Control}
Reinforcement learning and imitation learning have enabled humanoids to achieve expressive whole-body behaviors.
The seminal work DeepMimic~\cite{peng2018deepmimic} demonstrated that RL can track motion capture references with physical realism, inspiring benchmarks such as Humanoid-Gym~\cite{gu2024humanoid} and HumanoidBench~\cite{sferrazza2024humanoidbench}.
Motion retargeting methods address the embodiment gap: H2O~\cite{he2024learning} and OmniH2O~\cite{he2024omnih2o} enable real-time teleoperation, ExBody~\cite{cheng2024expressive, ji2024exbody2} extends imitation to expressive upper-body motions, and SONIC~\cite{luo2025sonic} scales motion tracking with large-scale datasets.
Generative approaches such as BeyondMimic~\cite{liao2025beyondmimic} and MaskedManipulator~\cite{tessler2025maskedmanipulator} leverage diffusion models to synthesize versatile motions from sparse goals.

Command-driven frameworks enable task-level control: AMO~\cite{li2025amo} employs motion optimization, HOMIE~\cite{ben2025homie} uses isomorphic teleoperation, and HOVER~\cite{he2025hover} consolidates locomotion, standing, and manipulation into a unified policy.
Hierarchical architectures separate planning from motor control: R$^2$S$^2$~\cite{zhang2025unleashing} and WholeBodyVLA~\cite{jiang2025wholebodyvla} encode commands as latent skill codes, while FALCON~\cite{zhang2025falcon} and VisualMimic~\cite{yin2025visualmimic} transmit trajectory-level commands with clearer geometric semantics.

\subsection{Spatial Consistency in Humanoid Loco-Manipulation}
Most existing methods reference commands to the robot base, assuming a stable coordinate frame~\cite{wang2025robust, wang2024arm, sandakalum2022motion}.
Wheeled mobile manipulators such as Mobile ALOHA~\cite{fu2024mobile} largely maintain this assumption, achieving bimanual coordination via imitation on stable bases.
However, legged humanoids must contend with cumulative drift and high-frequency oscillations during dynamic locomotion~\cite{li2025clone}, compounded by tight upper-lower body coupling~\cite{gu2025humanoid}.
DexMan~\cite{hsieh2025dexman} explores floating-base manipulation without explicitly modeling leg-arm coupling.
FALCON~\cite{zhang2025falcon} and RAMBO~\cite{cheng2025rambo} improve tracking through force adaptation and hybrid model-based RL, but still operate in body-centric coordinates.

In summary, imitation-based methods optimize joint-space errors rather than task-space precision~\cite{araujo2025retargeting} and require dense references~\cite{pan2025spider}, while command-driven methods operate in body-centric frames that cannot maintain world-frame consistency.
HiWET addresses these limitations through world-frame end-effector tracking with an explicit Cartesian interface, enabling active workspace reshaping through locomotion while ensuring dynamic stability via a modular low-level policy.

\section{Problem Formulation}
\label{sec:problem_formulation}

\begin{figure*}[htbp]
  \centering
  \includegraphics[width=0.95\linewidth]{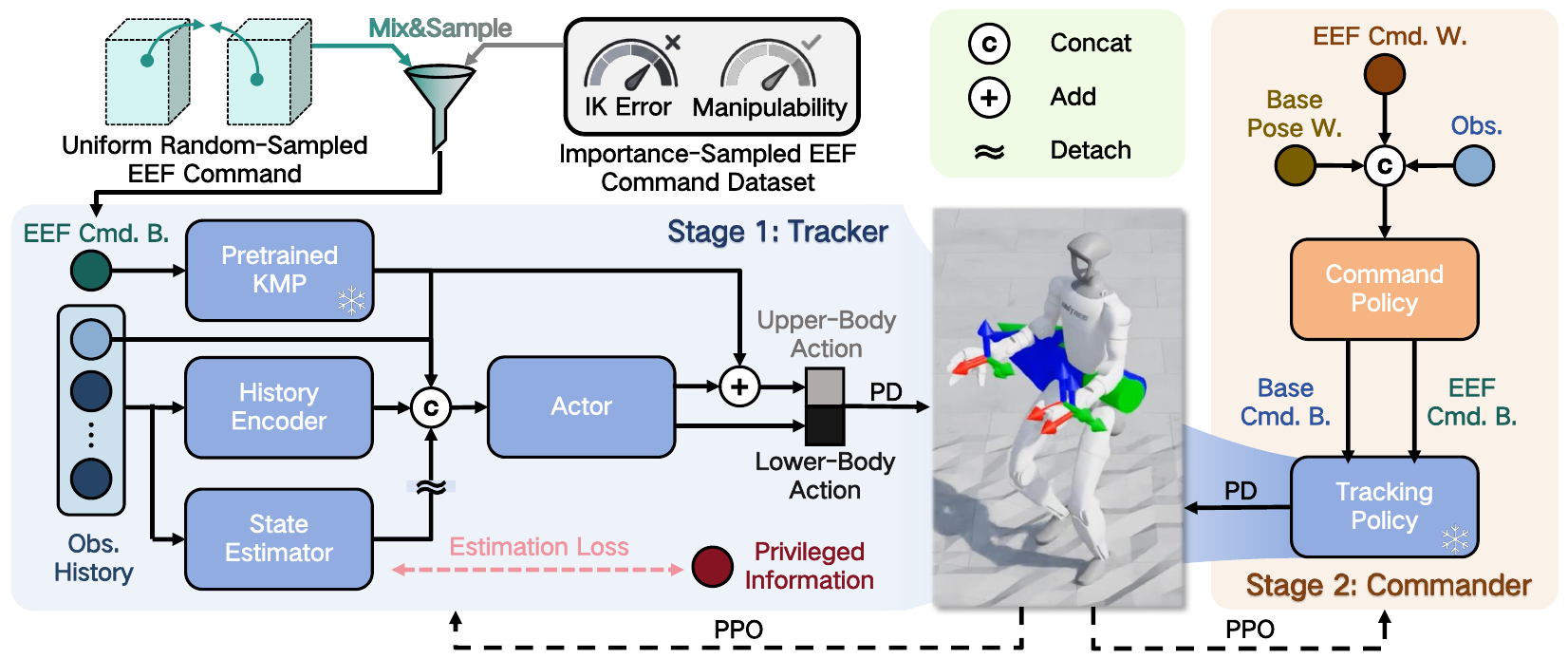}
  \caption{\textbf{HiWET architecture and two-stage training procedure.}
    \textbf{Stage 1: Tracker} (blue): The tracking policy learns to follow base-relative end-effector commands. Commands are sampled from a mixture of uniform random sampling and an importance-sampled dataset filtered by IK error and manipulability. A pretrained KMP provides upper-body kinematic references, which are refined by residual actions from the Actor. The History Encoder extracts temporal context, while the State Estimator reconstructs privileged information via an auxiliary estimation loss.
  \textbf{Stage 2: Commander} (orange): The command policy translates world-frame end-effector (EEF) targets (EEF Cmd. W.) and base pose into base-frame subgoals (Base Cmd. B. and EEF Cmd. B.).}
  \label{fig:method}
\end{figure*}

We propose a hierarchical reinforcement learning (HRL) framework that decouples global spatial reasoning from dynamic execution.
The overall architecture is illustrated in Fig.~\ref{fig:method}.
The \textit{world-frame command policy} functions as a task planner, interpreting global end-effector targets and generating local subgoals—specifically base velocity, body height, and base-relative hand poses.
The \textit{whole-body tracking policy} operates at a higher frequency to translate these subgoals into joint-space targets while ensuring physical feasibility and dynamic balance.
By formulating this interplay as a semi-Markov decision process (Semi-MDP), the hierarchy effectively reconciles long-horizon manipulation objectives with instantaneous stability constraints.

We model the humanoid control problem as a hierarchical reinforcement learning process
\begin{equation}
  \mathcal{M} = \langle \mathcal{S}, \mathcal{A}^H, \mathcal{A}^L, P, \mathcal{R}, \gamma \rangle,
\end{equation}
where $\mathcal{S}$ is the state space, $\mathcal{A}^H$ and $\mathcal{A}^L$ denote the high-level and low-level action spaces, $P$ is the transition dynamics, $\mathcal{R}$ is the reward function, and $\gamma \in (0,1)$ is the discount factor.

The hierarchy operates on two distinct timescales: the world-frame command policy $\pi^H$ runs every $K$ steps to update the subgoal command, while the whole-body tracking policy $\pi^L$ executes actions at every control step conditioned on the static high-level command.
Both policies are optimized to maximize the expected discounted return $J = \mathbb{E} [ \sum_{t=0}^{\infty} \gamma^t r_t ]$, where rewards are designed to balance global tracking precision with local dynamic stability.

\section{Whole-Body Tracking Policy}
\label{sec:low_level_policy}

The whole-body tracking policy translates high-level motion commands into joint-space targets for stable whole-body execution.
It operates at high frequency and interfaces directly with the PD controller.
The humanoid is decomposed into an upper-body subsystem (arms and waist) and a lower-body subsystem (legs).
This policy serves as a \textbf{standalone whole-body tracker} with a general-purpose command interface, capable of following arbitrary base velocity and end-effector subgoals independently of the high-level task planner.

\subsection{State and Action Representation}
\label{sec:low_level_state_action}

\subsubsection{State Space}
The whole-body tracking policy is conditioned on both the humanoid proprioceptive state $s_t$ and the structured command $\mathbf{u}_t$ received from the world-frame command policy.
The observation at time $t$ is defined as
\begin{equation}
  s_t =
  \big[
    \boldsymbol{\omega}_t,\;
    \mathbf{g}_t,\;
    \mathbf{q}_t,\;
    \dot{\mathbf{q}}_t,\;
    \mathbf{a}_{t-1}^L
  \big],
\end{equation}
where $\boldsymbol{\omega}_t \in \mathbb{R}^3$ is the base angular velocity, $\mathbf{g}_t \in \mathbb{R}^3$ is the gravity projection in the base frame, $\mathbf{q}_t$ and $\dot{\mathbf{q}}_t$ denote joint positions and velocities, and $\mathbf{a}_{t-1}^L$ is the previous action.

The command $\mathbf{u}_t$ transmits high-level spatial objectives to the low-level policy:
\begin{equation}
  \mathbf{u}_t =
  \big[
    \mathbf{v}_b^{des},\;
    h^{des},\;
    {}^{b}\mathbf{T}_{L}^{des},\; {}^{b}\mathbf{T}_{R}^{des},\;
    \alpha_t
  \big],
\end{equation}
where $\mathbf{v}_b^{des} = (v_x, v_y, \omega_{z})$ is the desired base velocity, $h^{des}$ is the target body height, ${}^{b}\mathbf{T}_{\{\cdot\}}^{des}$ denote the desired end-effector poses, and $\alpha_t$ is the waist regularization weight (uniformly sampled from $[0.1, 10]$ during training).
We use the superscript $\{\cdot\}^{des}$ to distinguish these local subgoals from global world-frame targets.

\subsubsection{Hybrid Action Space}
Given the proprioceptive observation $s_t$ and the command $\mathbf{u}_t$, the policy outputs joint position targets for all actuated joints:
\begin{equation}
  a_t^L = \mathbf{q}_t^{des} =
  \big[
    \mathbf{q}_{t,up}^{des},\;\mathbf{q}_{t,low}^{des}
  \big]
  \sim \pi^L(\mathbf{q}_t^{des} \mid s_t, \mathbf{u}_t),
\end{equation}
where $\mathbf{q}_{t,up}^{des}$ denotes the desired joint positions of the upper body and $\mathbf{q}_{t,low}^{des}$ denotes those of the lower body.

Instead of learning a monolithic policy, we adopt a hybrid action space that leverages kinematic structure.
For the upper body, the policy predicts a residual correction $\Delta \mathbf{q}_{t,up}$ which is added to the KMP reference $\hat{\mathbf{q}}_{t,up}$ to form the final command:
\begin{equation}
  \mathbf{q}_{t,up}^{des} = \hat{\mathbf{q}}_{t,up} + \Delta \mathbf{q}_{t,up}.
\end{equation}
This residual formulation enables the controller to make subtle dynamic adjustments for end-effector stability and accuracy while remaining within the valid manipulation manifold.
Conversely, the lower body is controlled via absolute joint targets $\mathbf{q}_{t,low}^{des}$ to maintain direct authority over gait generation and base propulsion.

\subsection{Network Architecture Overview}
\label{sec:network_architecture_overview}

The whole-body tracking policy employs a modular network architecture comprising three key components designed to address the challenges of partial observability and kinematic consistency (see Stage 1: Tracker in Fig.~\ref{fig:method}).

To ensure accurate end-effector tracking during motion and to provide a strong reference for learning, we introduce the Kinematic Manifold Prior (KMP) for the upper body.
The KMP predicts a kinematically consistent joint configuration given desired end-effector poses and a waist motion preference.
The input to the KMP is defined as
\begin{equation}
  \mathbf{z}_t =
  \big[
    {}^{b}\mathbf{T}_{L}^{des}, {}^{b}\mathbf{T}_{R}^{des},\;
    \alpha_t
  \big],
\end{equation}
where ${}^{b}\mathbf{T}_{\{\cdot\}}^{des}$ are the commanded end-effector poses, and $\alpha_t \in [0.1, 10]$ is a scalar waist regularization weight (the fixed weight for other upper-body joints is $0.2$).
This parameter serves as a hierarchical interface that modulates torso engagement.
Since humanoid stability is highly sensitive to the Center of Mass (CoM) position, explicitly allowing the global planner to control waist motion enables the framework to trade off between reachability and balance;
lower values of $\alpha_t$ encourage the KMP to utilize waist redundancy to extend the effective workspace, while larger values prioritize a stable CoM to maintain dynamic balance during locomotion.
The KMP produces a reference joint configuration for the upper body,
\begin{equation}
  \hat{\mathbf{q}}_{t,up} = f_{KMP}(\mathbf{z}_t),
\end{equation}
where $f_{KMP}$ is a neural approximation of the kinematic mapping for the arms and the waist.
The KMP is pretrained offline and frozen during policy optimization, enabling the policy to learn residual corrections rather than absolute joint targets.

\subsubsection{History Encoder}
To improve robustness under partial observability, we construct a history buffer of length $H=5$ containing tuples $(s_t, \mathbf{u}_t)$.
The History Encoder (a CNN) processes this buffer $\mathcal{H}_t = \{ (s_{t-i}, \mathbf{u}_{t-i}) \}_{i=0}^{H-1}$ and outputs a latent representation $\mathbf{e}_t$ encoding temporal dependencies.

\subsubsection{State Estimator}
The State Estimator (also a CNN) infers privileged state variables from the observation history.
It predicts base linear velocity and end-effector poses ${}^{b}\mathbf{T}_{L}, {}^{b}\mathbf{T}_{R}$—quantities essential for accurate tracking but not directly accessible from onboard sensors during deployment.
The estimator is trained jointly with the policy using a mean squared error loss.

\subsubsection{Actor and Critic Inputs}
The final input to the actor network is formed by concatenating the raw observation, the command, the privileged estimates from the State Estimator, the latent embedding from the History Encoder, and the kinematic reference from the frozen KMP:
\begin{equation}
  \mathbf{o}_t^{actor} =
  \big[
    s_t,\;
    \mathbf{u}_t,\;
    \hat{\mathbf{p}}_t,\;
    \mathbf{e}_t,\;
    \hat{\mathbf{q}}_{t,up}
  \big].
\end{equation}
The critic is trained with access to full privileged information to reduce variance in value estimation.
Its input is defined as
\begin{equation}
  \mathbf{o}_t^{critic} =
  \big[
    s_t,\;
    \mathbf{u}_t,\;
    \mathbf{p}_t,\;
    \mathbf{h}_t,\;
    \hat{\mathbf{q}}_{t,up}
  \big],
\end{equation}
where $\mathbf{p}_t$ includes the true base linear velocity and end-effector poses ${}^{b}\mathbf{T}_{L}, {}^{b}\mathbf{T}_{R}$, and $\mathbf{h}_t$ denotes terrain elevation measurements from the height map sensor.
Both actor and critic are optimized using Proximal Policy Optimization (PPO)~\cite{schulman2017proximal, rudin2022learning}.

\subsection{Command Importance Sampling for Policy Learning}
Uniform sampling in the Cartesian command space of the end-effectors leads to a significant proportion of kinematically infeasible or ill-conditioned targets, which typically lie near workspace boundaries and suffer from poor manipulability.
Training directly on such distributions degrades learning efficiency and stability.
To mitigate this, we curate a dataset using importance sampling (see top-left of Fig.~\ref{fig:method}).

To ensure that the policy focuses on functionally relevant and reachable configurations, we construct a prioritized end-effector command dataset.
Two metrics are used to curate this dataset.
First, the inverse kinematics reconstruction error is employed to prune unreachable poses for which the solver fails to converge.
Second, the manipulability index is used to weight sampling probability, favoring comfortable configurations that enable high-dexterity motion.

To preserve coverage of the full command space, the actual command fed into the policy during training is drawn from a mixture distribution,
\begin{equation}
  \mathbf{c} \sim \beta \, p_{uniform}(\mathbf{c}) + (1-\beta) \, p_{prior}(\mathbf{c}),
\end{equation}
where $\beta \in [0, 1]$ is the mixing weight, $p_{uniform}$ samples uniformly from the full Cartesian volume, and $p_{prior}$ samples from the curated dataset with added small perturbations.
This mixture ensures that the policy observes both realistic commands and boundary cases, improving robustness without sacrificing feasibility.

\subsection{Kinematic Manifold Prior Learning}
\label{sec:kmp_training}
\subsubsection{Dataset Generation}
Training data for the KMP are generated by solving constrained optimization-based inverse kinematics problems using PyRoki~\cite{kim2025pyroki}.
For each sampled end-effector command, an optimization problem is constructed to minimize Cartesian pose errors subject to joint limits and a regularization term on waist motion weighted by $\alpha$, which is uniformly sampled from $[0.1, 10]$.
This enables the prior to capture a spectrum of limb configurations ranging from rigid-torso reaching to whole-body lunges.
Approximately ten million samples are generated.

The initial end-effector command space is defined inside two symmetric Cartesian volumes around the torso center with full orientation coverage over the $SO(3)$ manifold.
However, not all commands in this space are kinematically feasible.
To curate the dataset, we apply importance sampling based on two metrics: the inverse kinematics reconstruction error, which prunes unreachable poses, and the manipulability index, which favors configurations that admit dexterous motion.
Specifically, fivefold oversampling is performed and filtered according to these criteria as detailed in the appendix.

\subsubsection{Network Architecture}
The KMP adopts a symmetric encoder-decoder architecture with residual connections.
It is based on a ResNet backbone~\cite{he2016deep} and incorporates layer normalization, dropout, and GELU activations to ensure stable training and generalization.

\subsection{Reward Formulation}

The reward function is designed to encourage accurate command tracking, stable locomotion, and precise manipulation, while preserving smooth and physically consistent motions.
In addition to standard tracking terms for base linear velocity, angular velocity, body height, and end-effector poses, several task-specific shaping rewards are introduced.

\subsubsection{KMP Reference Tracking}
To exploit the kinematic prior provided by the KMP, we include a reward that encourages the executed upper-body configuration to stay close to the reference.
Let $\hat{\mathbf{q}}_{t,up}$ denote the reference joint configuration from the KMP and $\mathbf{q}_{t,up}$ denote the current upper-body joint positions.
The tracking error is defined as
\begin{equation}
  r_{kmp,t} = \exp\left( - \frac{1}{N_{up} \sigma_{kmp}^2} \lVert \mathbf{q}_{t,up} - \hat{\mathbf{q}}_{t,up} \rVert_2^2 \right),
\end{equation}
where $\sigma_{kmp}$ controls the sensitivity.
A time-dependent scaling factor is applied after command resampling to avoid overly constraining early transient responses.

\subsubsection{Base Height Tracking with Knee Awareness}
To stabilize standing and squatting, we introduce a base height tracking term with asymmetric knee-aware weighting.
Let $h_t$ be the measured base height and $h_t^{des}$ the commanded height. The height error is weighted by knee joint margins: when above target, by flexion margin; when below, by extension margin:
\begin{equation}
  e_{h,t} =
  \begin{cases}
    (h_t - h_t^{des}) \cdot \bar{w}_{knee}^{flex} & \text{if } h_t > h_t^{des}, \\
    (h_t - h_t^{des}) \cdot \bar{w}_{knee}^{ext} & \text{if } h_t < h_t^{des},
  \end{cases}
\end{equation}
where $\bar{w}_{knee}^{flex}$ and $\bar{w}_{knee}^{ext}$ are the average margins for knee flexion and extension, respectively.
This term is activated only under zero velocity conditions to avoid interference with dynamic locomotion.

Additional regularization terms penalize excessive joint velocity, action rate, and torque variation to ensure smooth and hardware-friendly motions.
The complete reward formulation and weighting is provided in the appendix.

\section{World-Frame Command Policy}
\label{sec:high_level_policy}

The world-frame command policy $\pi^H$ functions as a global task planner, translating world-frame end-effector targets into dynamically feasible whole-body commands (see Stage 2: Commander in Fig.~\ref{fig:method}).
It reasons over the coupled nature of locomotion and manipulation, effectively enlarging the robot's operational workspace through active base transport.
During this training stage, the whole-body tracking policy from Stage 1 is frozen.

\subsection{State and Command Representation}
\label{sec:high_level_state_action}

\subsubsection{Observation and Task Command}
The world-frame command policy observes the proprioceptive state alongside global information and the task-level command $\mathbf{c}_t$.
The observation is defined as
\begin{equation}
  s_t^H =
  \big[
    s_t,\;
    {}^{w}\mathbf{T}_{b}^t,\;
    \mathbf{c}_t
  \big],
\end{equation}
where ${}^{w}\mathbf{T}_{b}^t \in SE(3)$ denotes the base pose in the world frame. The task command $\mathbf{c}_t$ is defined as
\begin{equation}
  \mathbf{c}_t =
  \Big[
    {}^{w}\mathbf{T}_{L}^{*},\; {}^{w}\mathbf{T}_{R}^{*},\;
    \mathbf{m}_t
  \Big],
\end{equation}
where ${}^{w}\mathbf{T}_{\{\cdot\}}^{*} \in SE(3)$ specify the desired end-effector poses in the absolute world frame, and $\mathbf{m}_t \in \{[1,0],[0,1],[1,1]\}$ is a binary mask that indicates which end-effector targets are active, enabling compatibility with both unilateral and bilateral manipulation tasks.

We employ an asymmetric actor-critic architecture: while the actor observes the above state, the critic is augmented with privileged information:
\begin{equation}
  \mathbf{o}_t^{H,\text{critic}} = \big[ s_t^{H},\; \mathbf{v}_b^{w,t},\; {}^{w}\mathbf{T}_{L}^t,\; {}^{w}\mathbf{T}_{R}^t \big],
\end{equation}
where $\mathbf{v}_b^w$ is the base linear velocity and ${}^{w}\mathbf{T}_{\{\cdot\}}^t$ are the current end-effector poses in the world frame.
The policy is optimized using PPO~\cite{schulman2017proximal, rudin2022learning}.

\subsubsection{Action Space}
The world-frame command policy outputs a structured motion command that serves as the input to the whole-body tracking policy:
\begin{equation}
  a_t^H = \mathbf{u}_t =
  \Big[
    \mathbf{v}_b^{des},\;
    h^{des},\;
    {}^{b}\mathbf{T}_{L}^{des},\; {}^{b}\mathbf{T}_{R}^{des},\;
    \alpha_t
  \Big]
  \sim \pi^H(\mathbf{u}_t \mid s_t^H).
\end{equation}
This abstraction allows the policy to focus on geometric consistency and reachability while delegating dynamic stability and contact management to the whole-body tracking policy.
The inclusion of $\alpha_t$ in the action space enables the high-level policy to actively regulate the robot's posture and CoM stability based on the global task context and locomotion state.

\subsection{Spatial Curriculum Strategy}
Training the world-frame command policy directly on large workspace ranges is challenging, as distant targets require long-horizon locomotion that complicates credit assignment.
We adopt a curriculum learning strategy that progressively expands the world-frame command range based on tracking performance.

Initially, world-frame targets ${}^{w}\mathbf{T}_{\{\cdot\}}^{*}$ are sampled within a small neighborhood around the robot's current base position.
As training progresses, we monitor the end-effector tracking error $e_{ee} = \| {}^{w}\mathbf{T}_{\{\cdot\}} - {}^{w}\mathbf{T}_{\{\cdot\}}^{*} \|$ and expand the sampling range when the error falls below a threshold.
Specifically, the planar command range $R$ is updated as:
\begin{equation}
  R_{k+1} = R_k + \Delta R \cdot \mathbb{1}[e_{ee} < \epsilon_{th}],
\end{equation}
where $\Delta R$ is the range increment and $\epsilon_{th}$ is the error threshold.
This curriculum ensures that the policy masters local coordination before tackling long-range workspace transport.

\subsection{Reward Formulation}
The high-level reward function is designed to reconcile absolute tracking precision with optimal base positioning.
\begin{itemize}
  \item \textbf{Workspace Optimization}: To maintain high manipulability, we penalize the planar distance between the robot base and the active end-effector targets.
    We further reward the alignment of both the base heading and the commanded velocity vector $\mathbf{v}_b^{des}$ with the vector pointing toward the target center, ensuring the policy actively ``steers'' the body to reshape the available workspace.
  \item \textbf{Precise End-effector Tracking}: Direct rewards are applied to minimize the pose error between ${}^{w}\mathbf{T}_{\{\cdot\}}$ and ${}^{w}\mathbf{T}_{\{\cdot\}}^{*}$ to enforce global spatial consistency.
  \item \textbf{Stability and Regularization}: To ensure hardware compatibility, we penalize action rates and high-frequency end-effector jitter.
    Additionally, when an arm is inactive (as indicated by $\mathbf{m}_t$), the policy is penalized for commanding deviations from a default neutral pose to prevent task-irrelevant limb drift.
\end{itemize}
A detailed description of reward components and weighting is provided in the appendix.
Through this hierarchical formulation, the world-frame command policy learns to coordinate locomotion and manipulation in the world frame, while delegating joint-space realization to the whole-body tracking policy.

\section{Experiments And Results}
The experiments aim to answer the following key questions.

\begin{enumerate}
  \item How do the constituent components of the whole-body tracking policy—kinematic prior, privileged estimation, and importance sampling—contribute to command tracking precision and stability?
  \item How does the world-frame command policy achieve spatial consistency and base positioning accuracy in long-horizon trajectory tracking tasks?
  \item How does the proposed KMP compare to optimization-based inverse kinematics solvers in terms of inference efficiency and reconstruction accuracy?
  \item Can the whole-body tracking policy successfully transfer to real hardware without additional fine-tuning?
\end{enumerate}

\subsection{Experimental Setup}
Our experimental platform is the Unitree G1 humanoid robot with 29 degrees of freedom (12 for legs, 14 for dual arms, and 3 for waist).
All policies are trained in Isaac Lab~\cite{mittal2025isaac} and deployed via zero-shot sim-to-real transfer.
Further details are provided in the Appendix.

\begin{table}[htbp]
  \centering
  \caption{Command Tracking Performance Comparison}
  \label{tab:tracking_comparison}
  \resizebox{\linewidth}{!}{
    \begin{tabular}{lcccc}
      \toprule
      Method & Lin. Vel. Error & Ang. Vel. Error & Height Error & EE Pos. Error \\
      & (m/s) & (rad/s) & (m) & (mm) \\
      \midrule
      HiWET             & 0.157 {\scriptsize $\pm$ 0.003} & 0.461 {\scriptsize $\pm$ 0.006} & 0.018 {\scriptsize $\pm$ 0.012} & 12.4 {\scriptsize $\pm$ 2.4} \\
      HiWET w/o IS      & 0.165 {\scriptsize $\pm$ 0.005} & 0.472 {\scriptsize $\pm$ 0.006} & 0.018 {\scriptsize $\pm$ 0.014} & 16.1 {\scriptsize $\pm$ 5.3} \\
      HiWET w/o State Est. & 0.169 {\scriptsize $\pm$ 0.003} & 0.459 {\scriptsize $\pm$ 0.004} & 0.018 {\scriptsize $\pm$ 0.016} & 23.0 {\scriptsize $\pm$ 7.2} \\
      HiWET w/o KMP & 0.149 {\scriptsize $\pm$ 0.004} & 0.423 {\scriptsize $\pm$ 0.005} & 0.015 {\scriptsize $\pm$ 0.010} & 25.2 {\scriptsize $\pm$ 12.8} \\
      HOMIE~\cite{ben2025homie} & 0.194 {\scriptsize $\pm$ 0.003} & 0.451 {\scriptsize $\pm$ 0.006} & 0.022 {\scriptsize $\pm$ 0.019} & - \\
      \bottomrule
    \end{tabular}
  }
\end{table}

\subsection{Low-Level Tracking Performance}

We evaluate the tracking performance of the low-level policy across various whole-body commands, measuring \textit{base linear velocity error}, \textit{angular velocity error}, \textit{body height error}, and \textit{hand Cartesian error}.
Commands are uniformly sampled within: linear velocity $v \in [-1, 1]$~m/s, angular velocity $\omega_z \in [-3, 3]$~rad/s, body height $h \in [0.3, 0.78]$~m, and end-effector poses within the maximal reachable Cartesian volume.
We compare against the state-of-the-art framework HOMIE~\cite{ben2025homie} and several ablation variants.
The quantitative results are summarized in Table~\ref{tab:tracking_comparison}.

\subsubsection{Comparison with Baselines}
HiWET outperforms the HOMIE baseline, reducing base linear velocity RMSE by $\sim$20\%.
Note that HOMIE receives joint position commands rather than end-effector poses, precluding direct end-effector comparison.
Our hierarchical formulation exposes locomotion-manipulation coupling, enabling more effective coordination while maintaining superior height consistency.

\subsubsection{Ablation Studies and Precision Improvements}
The ablation results highlight the critical contribution of each constituent component to the precision and robustness of bimanual tracking:
\begin{itemize}
  \item \textbf{HiWET w/o KMP}: Removing the KMP reference causes the most significant drop—hand error doubles (25.2~mm) with $5\times$ higher variance—confirming that learning Cartesian tasks without kinematic guidance is significantly more challenging.
  \item \textbf{HiWET w/o State Est.}: Without the state estimator, tracking degrades by nearly 10~mm, indicating that accurate end-effector feedback is essential for compensating locomotion-induced oscillations.
  \item \textbf{HiWET w/o IS}: Comparison with uniform sampling shows that focusing training on functional workspace regions yields more consistent bimanual coordination.
\end{itemize}
Additional ablation studies on reward components, including the knee-aware height tracking term, are provided in the Appendix.

Overall, the full HiWET framework achieves the highest precision (12.4~mm hand error) while maintaining the lowest variance, demonstrating its robustness as a general-purpose loco-manipulation tracker.

\begin{figure}[htbp]
  \centering
  \includegraphics[width=0.9\linewidth]{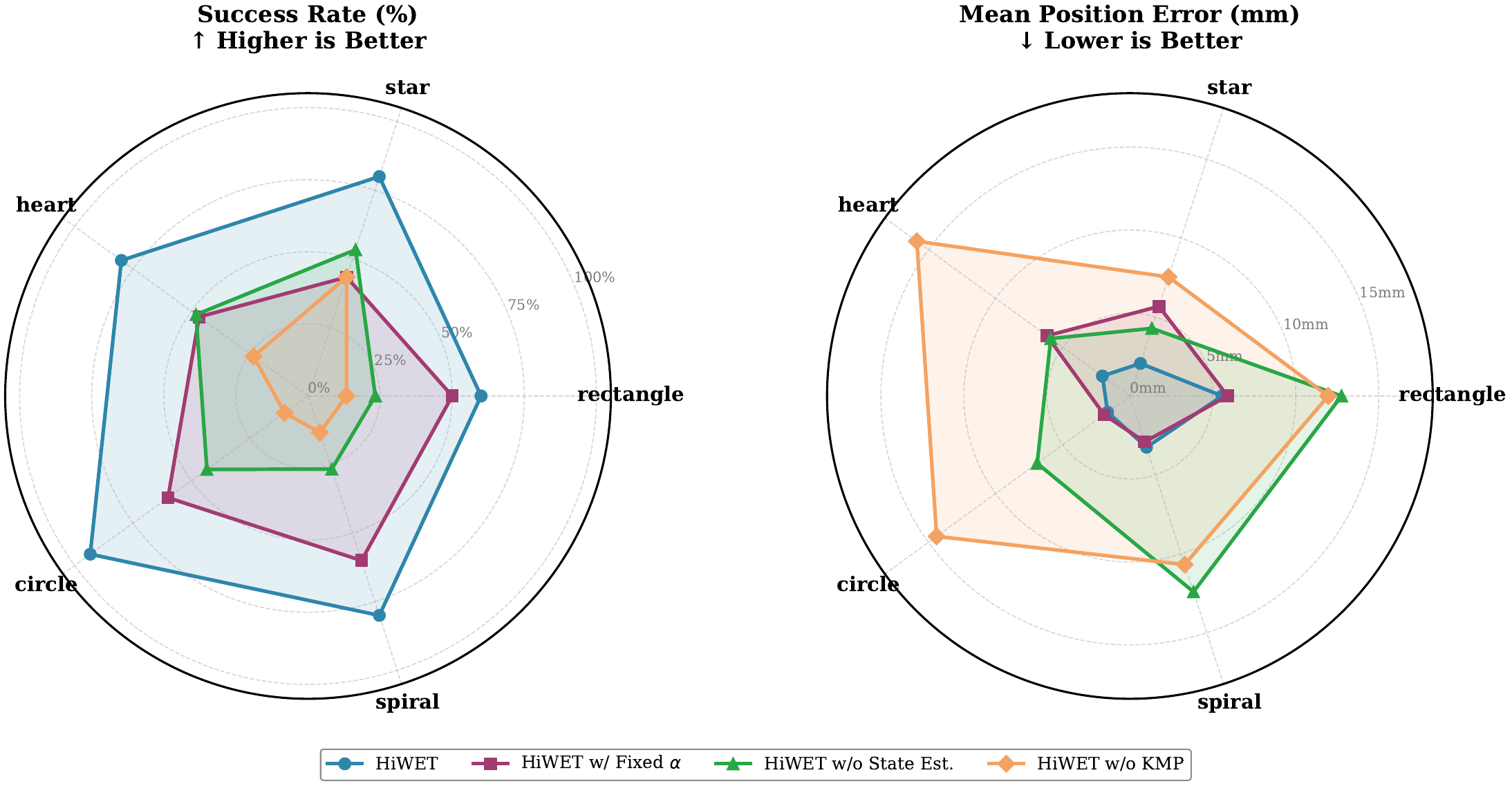}
  \caption{\textbf{Tracking success rate and mean position error across diverse geometric trajectories.}
  HiWET demonstrates improved performance and consistency compared to its ablated versions (w/ Fixed $\alpha$, w/o State Est., w/o KMP), particularly in complex tasks like tracing stars and hearts.}
  \label{fig:radar_chart}
\end{figure}

\subsection{World-Frame Navigation and Long-Horizon Tracking}

\subsubsection{Trajectory Tracking Evaluation}
We evaluate the capability of the high-level policy to perform spatially extended loco-manipulation tasks defined in absolute world coordinates, measuring \textit{tracking success rate} and \textit{mean end-effector position error}.
Geometric trajectories serve as standardized benchmarks to evaluate the controller's geometric precision and robustness across diverse motion profiles (e.g., sharp turns, velocity changes, symmetric/asymmetric movements).
Five distinct geometric trajectories—star, heart, circle, spiral, and rectangle—are randomly initialized at global positions within a range of $\pm 5$ m in both $x$ and $y$ axes relative to the robot's starting location.
This requires the robot to first navigate to the target region and then execute precise end-effector tracking while maintaining dynamic stability.

A trial is considered successful if the average end-effector tracking error remains below $20$ mm throughout the trajectory execution.
This criterion evaluates both the nav-manipulation coordination of the high-level policy and the disturbance rejection of the low-level policy.

\begin{figure}[htbp]
  \centering
  \includegraphics[width=\linewidth]{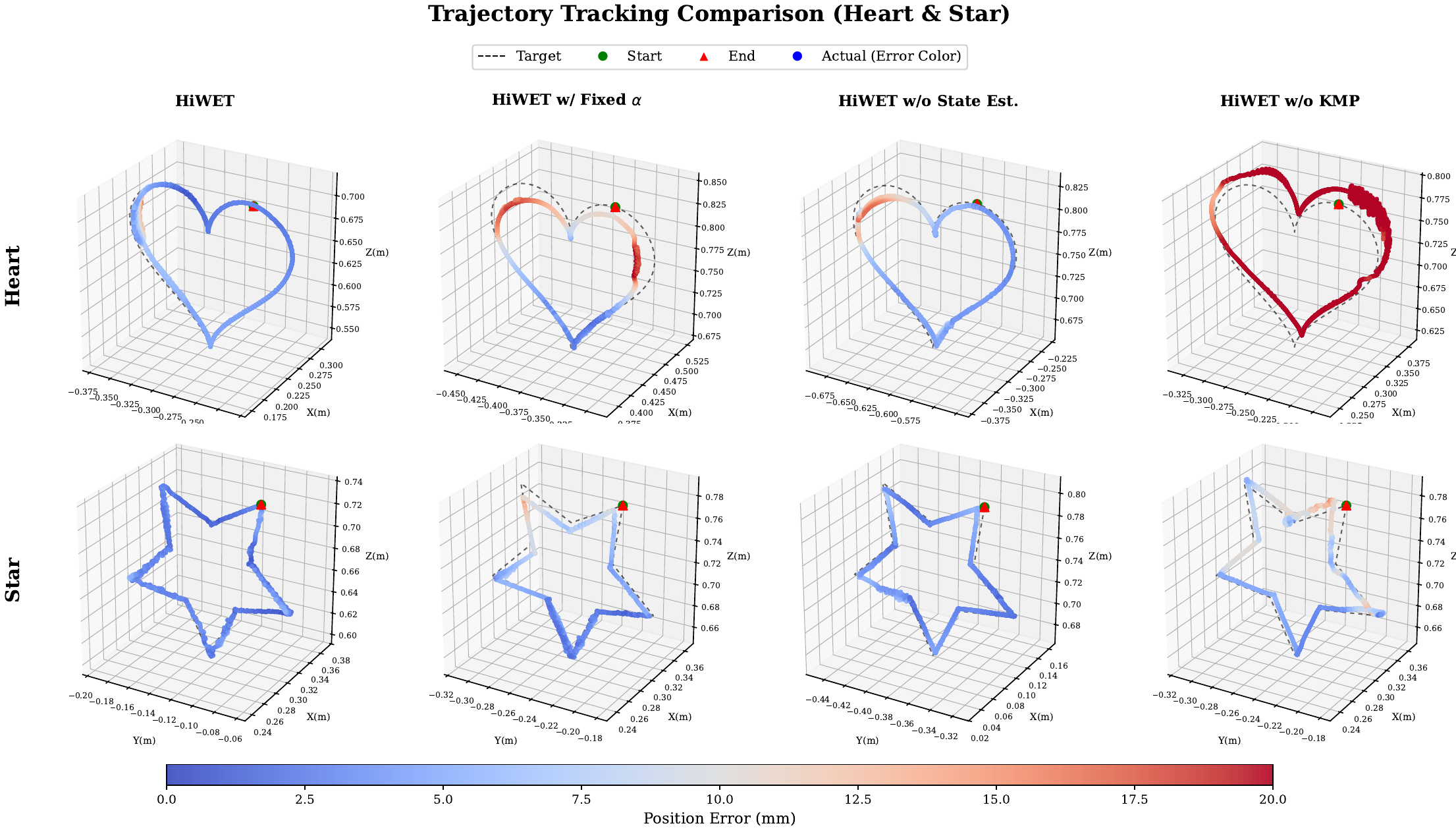}
  \caption{\textbf{Qualitative comparison of 3D world-frame trajectory tracking for heart and star shapes.}
    The color gradient indicates the instantaneous Cartesian position error (blue: $<2$ mm, red: $>20$ mm).
  HiWET (first column) maintains high spatial consistency, while fixing $\alpha=1.0$ (\textit{w/ Fixed $\alpha$}), removing privileged estimation (\textit{w/o State Est.}), or removing the kinematic prior (\textit{w/o KMP}) leads to varying degrees of deviation and oscillatory patterns. Additional geometric trajectories (circle, spiral, rectangle) are provided in the Appendix.}
  \label{fig:trajectory_spatial}
\end{figure}

As illustrated in Fig.~\ref{fig:radar_chart}, HiWET achieves high success rates and low tracking errors across all geometric patterns.
The qualitative comparison in Fig.~\ref{fig:trajectory_spatial} highlights heart and star trajectories, which require sharp changes in direction and coordinated end-effector movement.
HiWET (left column) consistently maintains errors below 5~mm, while the ablated variants exhibit significant performance degradation.
Specifically, the variant without the kinematic prior (\textit{w/o KMP}) shows severe trajectory distortions, as the high-level commands often drive the arms into kinematically ill-conditioned regions.
The removal of the privileged estimator (\textit{w/o State Est.}) leads to increased oscillations and tracking errors, highlighting the necessity of absolute state feedback for long-horizon spatial consistency.
The variant with fixed $\alpha=1.0$ (\textit{w/ Fixed $\alpha$}) exhibits degraded performance in tasks requiring large reaching motions, as the policy cannot exploit waist redundancy to control the CoM, thereby lacking the ability to dynamically trade off between reachability and stability.
Results for additional geometric trajectories (circle, spiral, rectangle) are provided in the Appendix.

\begin{figure}[htbp]
  \centering
  \includegraphics[width=\linewidth]{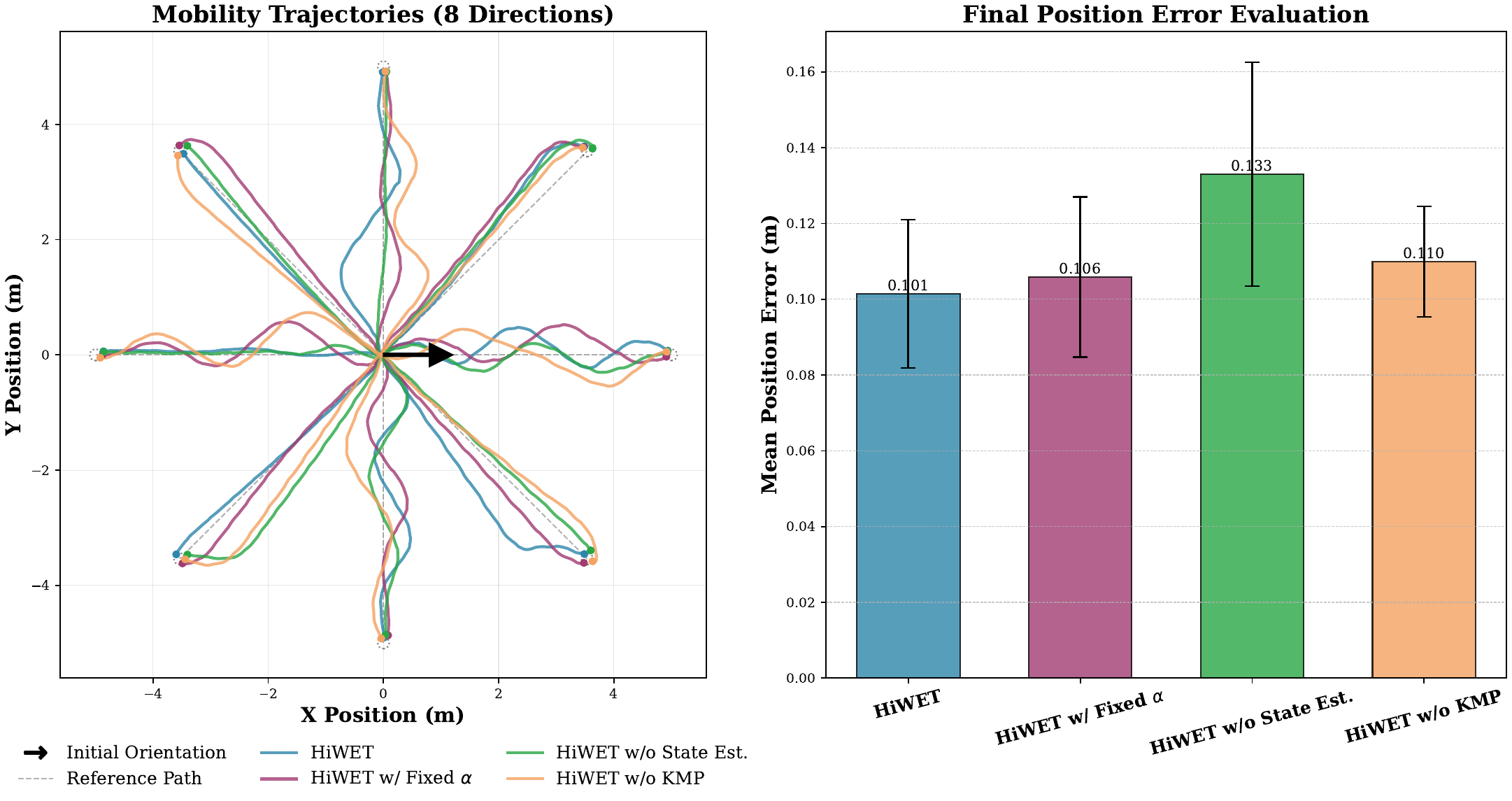}
  \caption{\textbf{Analysis of base mobility and positioning precision.} (Left) Top-view trajectories for 8-directional base repositioning tasks. (Right) Evaluation of final base positioning error in the $xy$-plane at the target location.
  HiWET provides the most stable trajectories and the highest positioning accuracy among all variants (w/ Fixed $\alpha$, w/o State Est., w/o KMP).}
  \label{fig:mobility_chart}
\end{figure}

\subsubsection{Base Mobility Assessment}
To isolate the robot's navigation performance, we evaluate its ability to reposition its support base in response to task-space requirements.
We initialize targets in eight cardinal and ordinal directions around the robot at a distance of 5~m.
Figure~\ref{fig:mobility_chart} (left) illustrates the top-view trajectories of the robot base for HiWET and its ablated variants.

HiWET exhibits the most direct and stable paths toward the targets, demonstrating efficient whole-body coordination where locomotion is actively guided by the global task objective.
In contrast, variants lacking high-fidelity feedback (\textit{w/o State Est.}), kinematic priors (\textit{w/o KMP}), or adaptive $\alpha$ modulation (\textit{w/ Fixed $\alpha$}) show oscillations or ``weaving'' behaviors, as the high-level policy struggles to reconcile conflicting base and arm objectives.

We further quantify the positioning precision by measuring the final distance between the robot base and the target starting point in the $xy$-plane upon task completion.
As shown in Fig.~\ref{fig:mobility_chart} (right), HiWET achieves the lowest mean error (0.101~m) and significantly less variance than other methods.
This high positioning accuracy is a direct consequence of our active whole-body coordination strategy, which enables the policy to treat the base as an active degree of freedom for workspace optimization rather than an independent mobile platform.

\begin{figure}[htbp]
  \centering
  \includegraphics[width=0.9\linewidth]{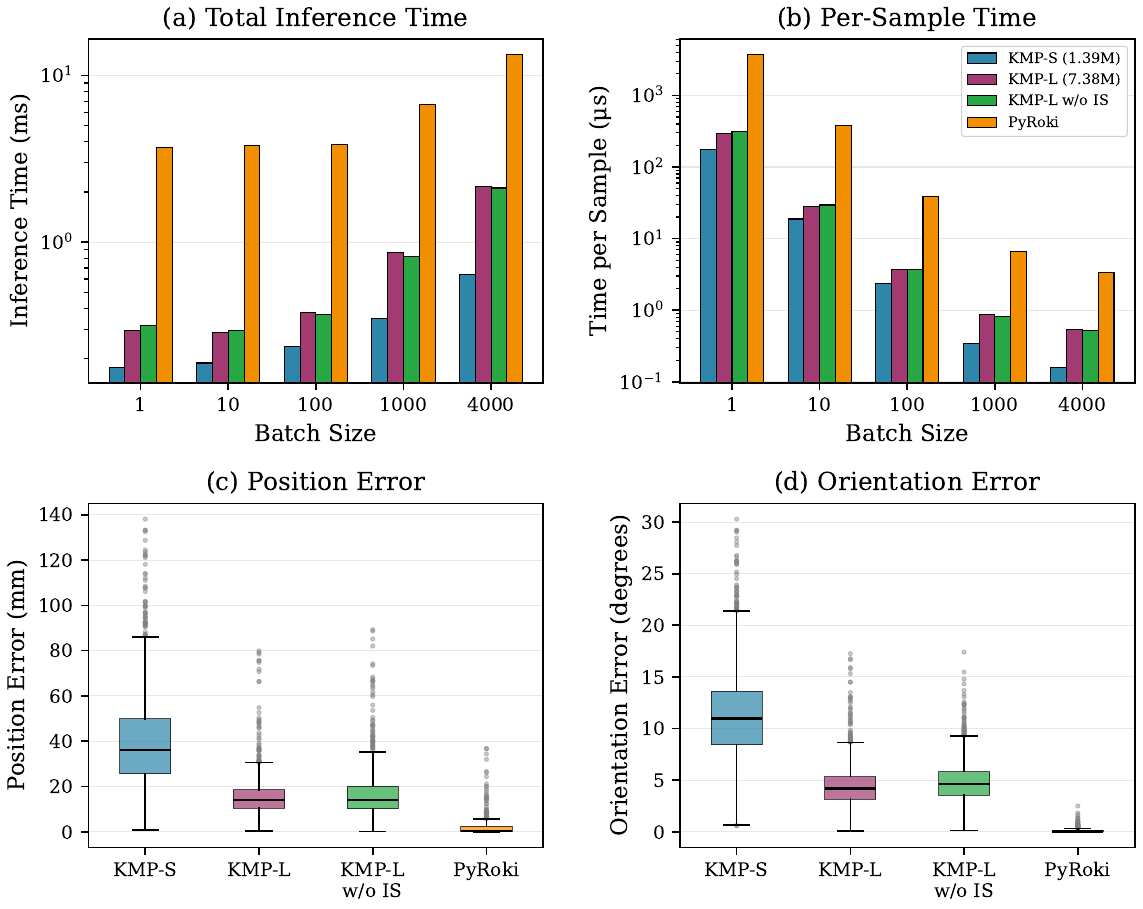}
  \caption{\textbf{Performance comparison between the proposed KMP variants and the optimization-based PyRoki solver.}
    (a)-(b) show that KMP variants achieve orders of magnitude speedup in inference latency, which is critical for real-time hierarchical control.
  (c)-(d) demonstrate that while optimization-based methods are near-exact, our KMP-L model trained with importance sampling (IS) provides sufficiently high precision for dual-arm coordination, serving as an efficient kinematic prior.}
  \label{fig:ik_performance}
\end{figure}
\subsection{Kinematic Manifold Prior Benchmarking}

To evaluate the proposed kinematic prior, we benchmark the performance of the KMP against an optimization-based solver, PyRoki~\cite{kim2025pyroki}, across two primary metrics: \textit{inference latency} for efficiency and \textit{Cartesian position/orientation error} for reconstruction accuracy.
We investigate two architectural variants: \textbf{KMP-S}, a lightweight 4-layer ResNet with smaller hidden dimensions, and \textbf{KMP-L}, a deeper 5-layer ResNet with larger hidden dimensions.
Furthermore, we examine the impact of training distributions by comparing models trained on our importance-sampled (IS) dataset against those trained on a uniform Cartesian cube (w/o IS) with an equivalent number of samples.

\subsubsection{Inference Efficiency}
As shown in Fig.~\ref{fig:ik_performance}(a)-(b), the KMP variants achieve significant computational speedups.
Compared to the PyRoki solver configured with 5 maximum iterations, KMP-L exhibits over $5\times$ speedup at a single-sample inference level.
This performance gap widens drastically under parallelized execution; at a batch size of 4000, KMP-L maintains millisecond-level latency.
This efficiency enables seamless integration into the reinforcement learning loop, providing high-frequency kinematic references without imposing a bottleneck on training throughput.

\subsubsection{Reconstruction Accuracy}
We evaluate the precision of the kinematic mapping using a test set derived from AMASS motion capture data~\cite{mahmood2019amass} retargeted to the G1 humanoid, which captures the natural spatial distribution of human end-effector usage.
Figures~\ref{fig:ik_performance}(c) and (d) summarize the Cartesian position and orientation errors.
While PyRoki remains the most precise due to its iterative nature, KMP-L achieves a median position error below 15~mm and orientation error below 5 degrees.
Notably, the KMP-L models trained with importance sampling (IS) significantly outperform their uniform counterparts (KMP-L w/o IS), demonstrating that focusing on functional and manipulable regions of the workspace is essential for high-fidelity bimanual coordination.
These results validate that KMP-L provides a sufficiently accurate and highly efficient kinematic prior for whole-body control.

\begin{figure}[htbp]
  \centering
  \includegraphics[width=0.85\linewidth]{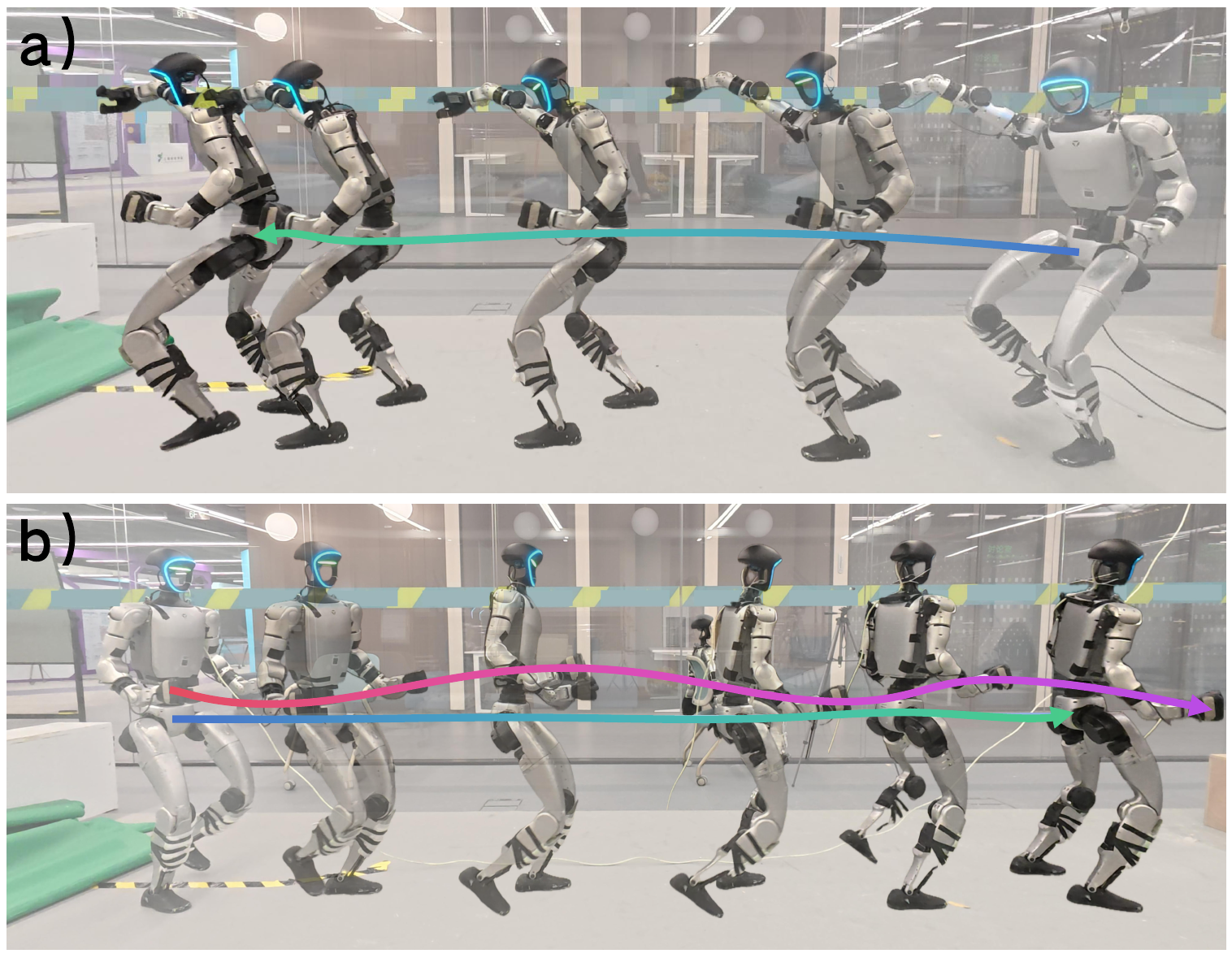}
  \caption{\textbf{Snapshots of real-world deployment on the Unitree G1 robot.}
    The low-level policy maintains stable locomotion while tracking diverse arm motions: (a) unilateral right-arm reaching, and (b) dynamic upper-body motion with the right hand tracing a circular trajectory (magenta curve indicates the right end-effector path).
  The system successfully rejects dynamic disturbances caused by rapid limb movements.}
  \label{fig:walk}
\end{figure}

\subsection{Real-world Hardware Experiments}
\subsubsection{Whole-body Tracking Policy}

To evaluate the transferability and robustness of the proposed framework, we deploy the low-level tracking policy on the physical Unitree G1 robot.
The policy is trained entirely in simulation and deployed via zero-shot transfer to the robot's onboard computer, operating at 50 Hz.

As illustrated in Fig.~\ref{fig:walk}, the robot demonstrates stable locomotion under varying and asymmetric limb configurations.
In Fig.~\ref{fig:walk}(a), the robot maintains a steady walking gait while the right arm is commanded to a high reaching pose.
In Fig.~\ref{fig:walk}(b), the robot successfully navigates while the right hand dynamically traces a circular trajectory, showcasing the policy's ability to handle rapid upper-body motions that induce significant center of mass (CoM) shifts and dynamic coupling between the limbs.
These experiments confirm that the low-level policy effectively manages the trade-off between task-space tracking and dynamic balance in real-world scenarios.

\begin{table}[htbp]
  \centering
  \caption{\textbf{Real-world end-effector tracking errors for circle and square trajectory tasks.}}
  \label{tab:real_world_tracking_error}
  \begin{tabular}{lcc}
    \toprule
    Method & Circle RMSE (m) & Square RMSE (m) \\
    \midrule
    HiWET                       & 0.012 {\scriptsize $\pm$ 0.005} & 0.015 {\scriptsize $\pm$ 0.007} \\
    HiWET w/ Fixed $\alpha$     & 0.018 {\scriptsize $\pm$ 0.008} & 0.019 {\scriptsize $\pm$ 0.009} \\
    HiWET w/o State Est.        & 0.024 {\scriptsize $\pm$ 0.011} & 0.028 {\scriptsize $\pm$ 0.011} \\
    HiWET w/o KMP               & 0.032 {\scriptsize $\pm$ 0.013} & 0.039 {\scriptsize $\pm$ 0.015} \\
    \bottomrule
  \end{tabular}
\end{table}

\subsubsection{World-Frame Command Policy}

To enable world-frame tracking, global pose is estimated using a head-mounted Livox Mid-360 LiDAR and IMU via Fast-LIO2~\cite{xu2022fast}, with base position updated at 10~Hz through forward kinematics.
In the circle and square trajectory tasks (Fig.~\ref{fig:intro}(e,f)), we compare HiWET against ablated variants over 10 repetitions.
As shown in Table~\ref{tab:real_world_tracking_error}, HiWET achieves the lowest tracking errors (12~mm for circles, 15~mm for squares).
The fixed $\alpha=1.0$ variant shows moderate degradation due to its inability to modulate torso engagement, while removing the state estimator or KMP leads to significantly larger errors.

\section{Conclusion And Limitations}
\label{sec:conclusion}

We presented HiWET, a hierarchical reinforcement learning framework for whole-body loco-manipulation.
By decoupling global task reasoning from dynamic execution, our approach effectively resolves the spatial and dynamic coupling between legged locomotion and precise manipulation.
The high-level planner operates in a world coordinate frame to ensure long-horizon consistency, while the low-level policy augmented by an efficient KMP provides the necessary dynamic stabilization.
Extensive simulation and real-world experiments validate the effectiveness of the proposed framework, achieving 12.4~mm mean tracking error in simulation and robust zero-shot sim-to-real transfer.

The current framework has \textbf{limitations}:
(1) world-frame tracking precision is bounded by LiDAR-based localization accuracy;
(2) evaluated trajectories are relatively small in scale;
(3) bimanual global tracking depends on target feasibility, and high-level experiments involve only single-arm tracking;
(4) contact-rich manipulation (e.g., grasping) remains unexplored.
Future work will integrate visual perception and force control for dexterous, contact-aware manipulation.



\bibliographystyle{plainnat}
\bibliography{references}

\end{document}